\documentclass[11pt,a4paper]{article}
\usepackage{authblk}
\usepackage[hyperref]{acl2019}
\usepackage{times}
\usepackage{latexsym}
\usepackage{graphicx}
\usepackage{url}
\usepackage{amsmath}
\usepackage{amssymb}
\usepackage{multirow}
\usepackage{tablefootnote}
\usepackage{multicol}

\aclfinalcopy % Uncomment this line for the final submission
\def\aclpaperid{***} %  Enter the acl Paper ID here

\title{Multi-Task Learning for Coherence Modeling}

\author[ ]{\textbf{Youmna Farag}}
\author[ ]{\textbf{Helen Yannakoudakis}}
\affil[ ]{Department of Computer Science and Technology}
\affil[ ]{The ALTA Institute}
\affil[ ]{University of Cambridge}
\affil[ ]{United Kingdom}
\affil[ ]{\tt \{youmna.farag,helen.yannakoudakis\}@cl.cam.ac.uk}

\date{}

\begin{document}
\maketitle
\begin{abstract}
We address the task of assessing discourse coherence, an aspect of text quality that is essential for many NLP tasks, such as summarization and language assessment. We propose a hierarchical neural network trained in a multi-task fashion that learns to predict a document-level coherence score (at the network's top layers) along with word-level grammatical roles (at the bottom layers), taking advantage of inductive transfer between the two tasks.  
We assess the extent to which our framework generalizes to different domains and prediction tasks, and demonstrate its effectiveness not only on standard binary evaluation coherence tasks, but also on real-world tasks involving the prediction of varying degrees of coherence, achieving a new state of the art. 
\end{abstract}

\section{Introduction}
Discourse coherence refers to the way textual units relate to one another and form a coherent whole. Coherence is an important aspect of text quality and therefore its modeling is essential in many NLP applications, including summarization \cite{Barzilay2002,Daraksha2016}, question-answering \cite{verberne2007evaluating}, question generation  \cite{desai2018generating}, and language assessment \cite{Burstein2010,Somasundaran2014,Farag2018}. A large body of work has investigated models for the assessment of inter-sentential coherence, that is, assessment in terms of transitions between adjacent sentences \cite{Barzilay2008,y2012modeling,Guinaudeau2013,Dat2017,Joty2018}. The properties of text that result in inter-sentential connectedness have been translated into a number of computational models -- some of the most prominent ones include the entity-based approaches, inspired by Centering Theory \cite{Grosz1995} and proposed in the pioneering work of \citet{Barzilay:2005,Barzilay2008}. Such approaches model local coherence in terms of entity transitions between adjacent sentences, where entities are represented by their syntactic role in the sentence (e.g., subject, object). 

Current state-of-the-art deep learning adaptations of the entity-based framework involve the use of Convolutional Neural Networks (CNNs) over an entity-based representation of text to discriminate between a coherent document and its incoherent variants containing a random reordering of the document's sentences \cite{Dat2017}; as well as lexicalized counterparts of such models that further incorporate lexical information regarding the entities, thereby distinguishing between different entities \cite{Joty2018}. 

In contrast to existing approaches, we propose a more generalized framework that allows neural models to encode information about the types of grammatical roles all words in a sentence participate in, rather than focusing only on the roles of entities within a sentence. Inspired by recent advances in Multi-Task Learning (MTL) \cite{rei2017auxiliary,sanh2018hierarchical}, 
we propose a simple, yet effective hierarchical model trained in a multi-task fashion that learns to perform two tasks: scoring a document's discourse coherence and predicting the type of grammatical role (GR) of a dependent with its head.  
We take advantage of inductive transfer between these tasks by giving a supervision signal at the bottom layers of a network with respect to the types of GRs, and a supervision signal at the top layers with respect to document-level coherence. 

Our contributions are four-fold: (1) We propose a MTL approach to coherence assessment and compare it against a number of baselines. We experimentally demonstrate that such a framework allows us to exploit more effectively the inter-dependencies between the two prediction tasks and achieve state-of-the-art results in predicting document-level coherence; (2) We assess the extent to which the information encoded in the network generalizes to different domains and prediction tasks, and demonstrate the effectiveness of our approach not only on standard binary evaluation tasks on the Wall Street Journal (WSJ), but also on more realistic tasks involving the prediction of varying degrees of coherence in people's everyday writing; (3) In contrast to existing work that has only investigated the impact of a specific set of grammatical roles (i.e., subject and object) on coherence, we instead investigate a large set of GR types, and train the model to predict the type of role dependents participate in. 
This allows the network to learn more generic patterns of language and composition, and a much richer set of representations than those induced by current approaches. In turn, this can be better exploited at the top layers of the network for predicting document-level coherence; (4) Finally, and contrary to previous work, our model does not rely on the availability of external linguistic tools at testing time as it directly learns to predict the GR types.

%Figure 1
%%%%%%%%%%%%%%%%%%%%%%%%%%%%%%%
\begin{figure*}[t]
\centering
\includegraphics[scale=0.4]{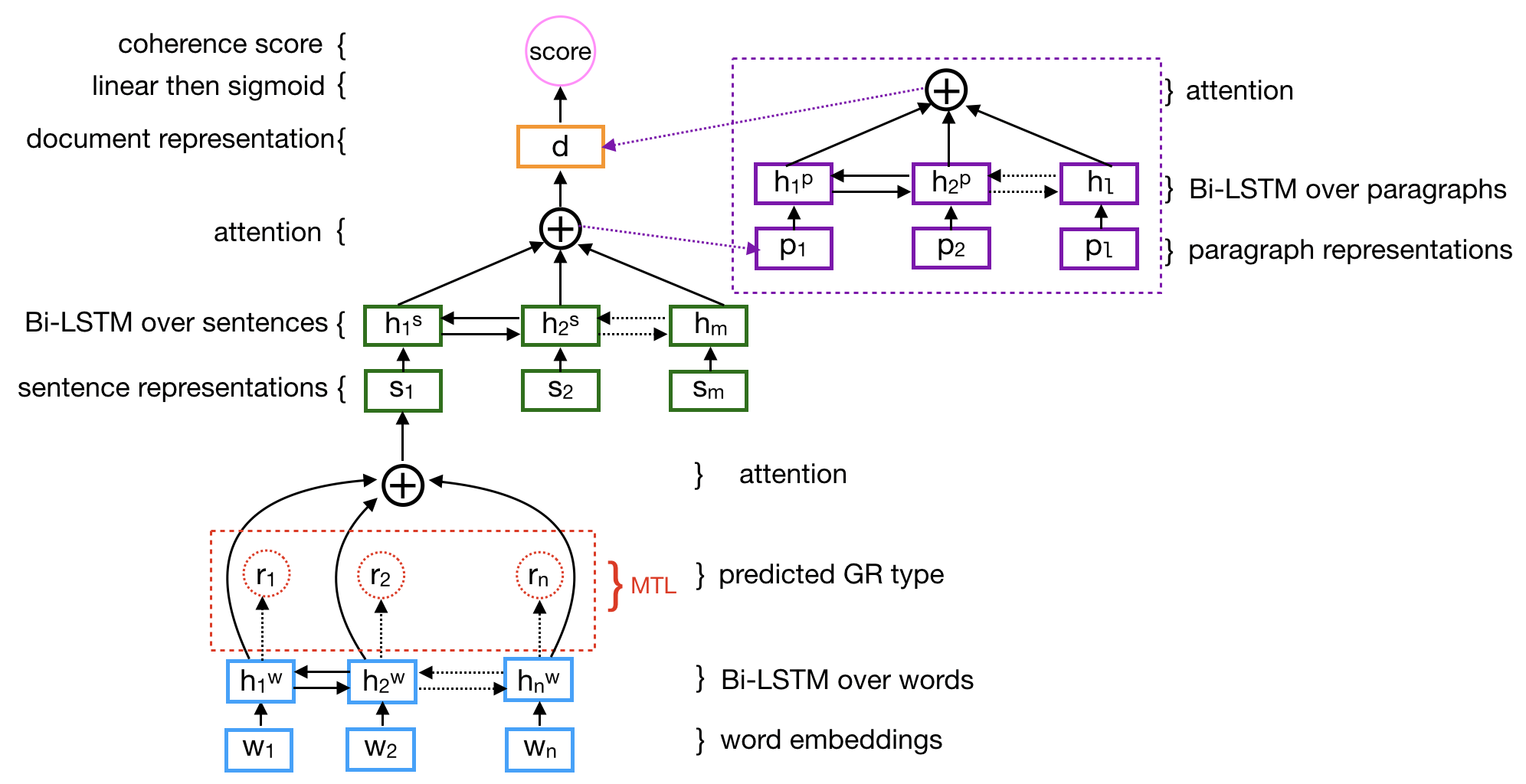}
\caption{The hierarchical architecture of the STL and MTL models. The dotted red box is specific to the MTL framework. The dotted purple box is applied if the document contains paragraph boundaries (which is the case for the Grammarly Corpus in Section \ref{data}) in order to create paragraph representations prior to the document one.}
\label{figure1}
\end{figure*}
%%%%%%%%%%%%%%%%%%%%%%%%%%%%%%%
\section{Related Work}
Several studies have proposed frameworks for modeling the textual properties that coherent texts exhibit. A popular approach is one based on the entity-grid (egrid) representation of texts, proposed by \citet{Barzilay:2005,Barzilay2008} and inspired by Centering Theory \cite{Grosz1995}. In the egrid model, texts are represented as matrices of entities (columns) and sentences (rows). Entities in the matrix are represented by their grammatical role (i.e., subject, object, neither), and entity transitions across sentences are used as features for coherence assessment. A large body of work has utilized and extended the egrid approach \cite{Elsner2008,Burstein2010,Elsner2011,Guinaudeau2013}. Other features have also been leveraged, such as syntactic patterns \cite{Louis2012} and discourse relations \cite{Lin2011,Feng2014}. 
Deep learning architectures have also been successfully applied to the task of coherence scoring, achieving state-of-the-art results \cite{Li2017,Logeswaran2018,Cui2018}. 
Some have exploited egrid features in a CNN model aimed at capturing long range entity transitions \cite{Dat2017,Joty2018}; further details are provided in Section~\ref{mandbase}.

Traditionally, coherence evaluation has been treated as a binary task, where a model is trained to distinguish between a coherent document and its incoherent counterparts created by randomly shuffling the sentences it contains. The news domain has been a popular source of well-written, coherent texts. Among the popular datasets are articles about \texttt{EARTHQUAKES} and \texttt{AIRPLANES} accidents \cite{Barzilay2008,Guinaudeau2013,Li2017} and the Wall Street Journal (WSJ) portion of the Penn Treebank \cite{Elsner2008, Lin2011, Dat2017}. \citet{Elsner2008} argue that the WSJ documents are normal informative articles, whereas the \texttt{AIRPLANES} and \texttt{EARTHQUAKES} ones have a more constrained style.

\section{Approach}
\label{stl_mtl}
\subsection{Neural Single-Task Learning (STL)}
\label{lex}
Our baseline model, shown in Figure~\ref{figure1}, performs the single task of predicting an overall coherence score via a hierarchical model based on a Long Short-Term Memory (LSTM) network \cite{Hochreiter1997}. A document is composed of a sequence of sentences $\{s_1, s_2, ..., s_m\}$ and, in turn, each sentence consists of a sequence of words $\{w_1, w_2, ..., w_n\}$. The input words are initialized with vectors from a pre-trained embedding space. A bidirectional LSTM (Bi-LSTM) is applied to the words in each sentence to get contextualized representations, and the output vectors from both directions are concatenated:
\begin{equation}
\label{eq1}
\begin{split}
\overrightarrow{h_t^{w}} &= LSTM(w_t,\overrightarrow{h_{t-1}^{w}}) \\ 
\overleftarrow{h_t^{w}} &= LSTM(w_t,\overleftarrow{h_{t+1}^{w}}) \\
h_t^{w} &= [\overrightarrow{h_{t}^{w}},\overleftarrow{h_{t}^{w}}]
\end{split}
\end{equation}
To compose a sentence representation $s$, the hidden states  $\{h_1^{w}, ...,h_n^{w}\}$ of its words are combined with an attention mechanism: 
\begin{equation}
\label{eq2}
\begin{split}
u_t^{w} &= tanh(W^{w} h_t^{w})  \\ 
a^{w}_{t} &= \frac{exp(v^{w} u_t^{w} )}{\sum_{t}{exp(v^{w} u_t^{w} )}} \\
s &= \sum_{t}{a^{w}_{t} h_t^{w}}
\end{split}
\end{equation}
where $W^{w}$ and $v^{w}$ are learnable parameters. Attention allows the model to focus on the salient words for coherence and build better sentence representations.

Constructing a document representation $d$ is similar to the sentence one -- a second Bi-LSTM is utilized over sentences $\{s_1, s_2, ..., s_m\}$ to generate contextually rich sentence representations:
\begin{equation}
\label{eq3}
\begin{split}
\overrightarrow{h_i^{s}} &= LSTM(s_i,\overrightarrow{h_{i-1}^{s}}) \\ 
\overleftarrow{h_i^{s}} &= LSTM(s_i,\overleftarrow{h_{i+1}^{s}}) \\
h_i^{s} &= [\overrightarrow{h_{i}^{s}},\overleftarrow{h_{i}^{s}}]
\end{split}
\end{equation}
Subsequently, attention is applied over the sentence embeddings $\{h_1^{s}, ...,h_m^{s}\}$ to allow the model to focus on sentences that contribute highly to the overall coherence of the document: 
\begin{equation}
\label{eq4}
\begin{split}
u_i^{s} &= tanh(W^{s} h_i^{s}) \\
a^{s}_{i} &= \frac{exp(v^{s} u_i^{s} )}{\sum_{i}{exp(v^{s} u_i^{s} )}} \\
d &= \sum_{i}{a^{s}_{t} h_i^{s}}
\end{split}
\end{equation}
where $W^{s}$ and $v^{s}$ are trainable weights in the network.
If a document consists of paragraphs $\{p_1, p_2, ..., p_l\}$, a third Bi-LSTM is stacked over the sentence vectors and the output is aggregated with another attention layer to compose the document vector $d$.

Finally, the coherence score of a document is predicted by applying a linear transformation to the vector $d$ followed by a sigmoid operation to bound the score in $[0,1]$:
\begin{equation}
\label{eq5}
\hat{y} =  \sigma(W^d \; d)
\end{equation}
where $W^d \in \mathbb{R}^{dim}$ is the linear function weight and $dim$ represents the dimensionality of the document vector. In a binary classification task, where the document is labeled as either coherent or incoherent, the model predicts one value for $\hat{y} \in [0,1]$. In a multiclass classification setting where there are multiple classes $y \in C$ representing various degrees of coherence, a document is labeled with a one-hot vector with length $|C|$ with a value of $1$ in the index of the correct class and $0$ everywhere else. The model predicts $|C|$ scores, using Equation \ref{eq5} with $W^d \in \mathbb{R}^{dim \times |C|}$, and learns to maximize the value corresponding to the gold label.

For the binary task, the network's parameters are optimized to minimize the negative log-likelihood of the document's ground-truth label $y$, given the network’s prediction $\hat{y}$:  
\begin{equation}
\label{eq6}
L_1 = -y \; log(\hat{y}) - (1 - y) log(1 - \hat{y})
\end{equation}
For the multiclass task, we use mean squared error to minimize the discrepancy between the one-hot gold vector and the estimated one: 
\begin{equation}
\label{eq7}
L_1 = \frac{1}{|C|}\sum_{j=1}^{|C|}{(y_j - \hat{y_j})^2}
\end{equation}
An alternative approach to the multiclass problem is to apply a softmax over the predictions instead of a sigmoid, and minimize the categorical cross entropy; however, initial experiments on the development set showed that our formation yields better results. 

\subsection{Neural Multi-Task Learning (MTL)}
\label{mtl}
The model described in \ref{lex} performs the single task of predicting a coherence score for a text; all model parameters are tuned to minimize the loss ($L_1$) in Equation \ref{eq6} or \ref{eq7} (depending on whether we are optimizing for a binary or a multiclass classification task respectively). 
We extend this model to a MTL framework by training it to optimize a secondary objective at the bottom layers of the network, along with the main one ($L_1$). Specifically, the model is trained to predict a document-level score along with word-level labels indicating the (predicted) GR type of dependents in the document.\footnote{{We make our code publicly available at \url{https://github.com/Youmna-H/coherence_mtl}}} 
The GRs are based on a predefined set $R$, generated from a dependency parser on the training set (Section \ref{expsetup}). 
The set includes the types of GRs in which a word is a dependent (e.g., \textit{nsubj}, \textit{amod}, \textit{xcomp}, \textit{iobj}), and each type $r\in R$ is treated as a class (for the `root' word, the type is \textit{root}). 
In order to predict a probability distribution over $R$ given a word representation $h_t$ (Equation \ref{eq1}), a linear operation normalized by a softmax function is applied: 
\begin{equation}
\label{eq8}
P(y^r_t|h^w_t) = softmax(W^r h^w_t)
\end{equation}
The secondary objective and the word-level loss is defined as the categorical cross-entropy, i.e., the negative log-probability of the correct labels:
\begin{equation}
\label{eq9}
L_2 = -\sum_t{\sum_r{y^r_t log{P(y^r_t|h^w_t)}}}
\end{equation}
Both the main ($L_1$) and secondary ($L_2$) objectives are optimized jointly ($L_{total}$), but with different weights to indicate the importance of each of these tasks during training:
\begin{equation}
\label{eq10}
L_{total} = \alpha L_1 + \beta L_2
\end{equation}
where $\alpha, \beta \in [0,1]$ are the loss weight hyperparameters. Figure \ref{figure1} (red-dotted box) presents the complete MTL framework. MTL allows us to take advantage of inductive transfer between these tasks and learn a rich set of representations at the bottom layers that can be exploited by the top layers of the network for predicting a document-level coherence score. 

Current state-of-the-art approaches utilizing the entity-based framework \cite{Joty2018} focus solely on the subject and object types. To further assess the impact of our extended set of GR types, we re-train the same MTL model but now only utilize subject ($S$) and object ($O$) GR types as our secondary training signal. Following the current entity-based approaches, all other types are mapped to $X$, to represent `other' roles;
specifically, $R=\{S,O,X\}$. We refer to this baseline model as {MTL\textsubscript{sox}}.   

 \section{Experiments}
 \subsection{Data and Evaluation Metrics}
 \label{data}
 \noindent{\bf Synthetic Data.} The Wall Street Journal (WSJ) portion of the Penn Treebank \cite{Elsner2008, Lin2011, Dat2017} is one of the most popular datasets for (binary) coherence assessment, given its size and the nature of the texts it contains; i.e. long articles not constrained in style \cite{Elsner2008,Dat2017}. Following previous work \cite{Dat2017}, we also use the WSJ and specifically sections $00-13$ for training and $14-24$ for testing (documents consisting of one sentence are removed). We create $20$ permutations per document, making sure to exclude duplicates or versions that happen to have the same ordering of sentences as the original article. Table \ref{table1} presents the data statistics.

 %Table 1
%%%%%%%%%%%%%%%%%%%%%%%%%%%%%%%
\begin{table}[]
\centering
\scalebox{0.9}{
\begin{tabular}{cccc}
\hline
      & \#Docs & \#Synthetic Docs & Avg \#Sents \\ \hline
 \multicolumn{1}{c|}{Train} & 1,376   & 25,767            & 21.0         \\ \hline
 \multicolumn{1}{c|}{Test}  & 1,090   & 20,766            & 21.9         \\ \hline
\end{tabular}}
\caption{Statistics for the WSJ data. \#Docs represents the number of original articles and \#Synthetic Docs the number of original articles + their permuted versions.}
\label{table1}
\end{table}
%%%%%%%%%%%%%%%%%%%%%%%%%%%%%%%

%Table 2
%%%%%%%%%%%%%%%%%%%%%%%%%%%%%%%
\begin{table}[]
\centering
\scalebox{0.9}{
\begin{tabular}{cccc}
\hline
                         &       & \#Docs & Avg \#Sents \\ \hline
\multicolumn{1}{c|}{\multirow{2}{*}{Yahoo}}   & \multicolumn{1}{c|}{Train} & 1000    & 7.5        \\ \cline{2-4} 
\multicolumn{1}{c|}{}                          & \multicolumn{1}{c|}{Test}  & 200     & 7.5        \\ \hline
\multicolumn{1}{c|}{\multirow{2}{*}{Clinton}}  & \multicolumn{1}{c|}{Train} & 1000    & 6.6        \\ \cline{2-4} 
\multicolumn{1}{c|}{}                          &  \multicolumn{1}{c|}{Test}  & 200     & 6.6        \\ \hline
\multicolumn{1}{c|}{\multirow{2}{*}{Enron}}   & \multicolumn{1}{c|}{Train} & 1000    & 7.7        \\ \cline{2-4} 
 \multicolumn{1}{c|}{}                         &  \multicolumn{1}{c|}{Test}  & 200     & 7.8        \\ \hline
\end{tabular}
}
\caption{Statistics for the GCDC.}
\label{table2}
\end{table}
%%%%%%%%%%%%%%%%%%%%%%%%%%%%%%%

To evaluate model performance on this dataset, we again follow previous work \cite{Barzilay2008,Dat2017} and calculate pairwise ranking accuracy (PRA) between an original text and its $20$ permuted counterparts. Specifically, PRA calculates the fraction of correct pairwise rankings in the test data (i.e., a coherent/original text should be ranked higher than its permuted counterpart). Following \citet{Farag2018}, we also report the total pairwise ranking accuracy (TPRA) that extends PRA to comparing each original text to all permuted texts in the test set rather than only its own set of permuted counterparts. \\
\vspace{-0.4cm}  

\noindent{\bf Realistic Data.}\label{rwd}
The Grammarly Corpus of Discourse Coherence (GCDC) is a newly-released dataset containing emails and reviews written with varying degrees of proficiency and care \cite{Lai2018}.\footnote{\url{https://github.com/aylai/GCDC-corpus}} In addition to the WSJ, we employ this dataset in order to assess the effectiveness of our coherence model for tasks involving the prediction of varying degrees of coherence in people's everyday writing. Specifically, the dataset contains texts from four domains: \textbf{Yahoo} online forum posts, emails from Hillary \textbf{Clinton}'s office, emails from \textbf{Enron} and \textbf{Yelp} business reviews. As some of the reviews from the latter were subsequently removed by Yelp, we evaluate our model on each of the first three domains (Table \ref{table2}). 

Annotators were instructed to rate each document with a score $\in \{1, 2, 3\}$, representing \textit{low}, \textit{medium} and \textit{high} levels of coherence respectively. For our experiments, we use the consensus rating of the expert scores as calculated by \citet{Lai2018}, and train the models to maximize the probability of the gold class within a multiclass classification framework (see Section \ref{stl_mtl}). The gold label distribution is as follows: Yahoo $44.8\%$ low, $17.9\%$ medium, $37.25\%$ high; Clinton $27.8\%$ low, $20.3\%$ medium, $51.8\%$ high; Enron $30\%$ low, $20.3\%$ medium, $49.6\%$ high. 
To evaluate model performance, we use three-way classification accuracy.

\subsection{Models and Baselines}
\label{mandbase}
\noindent{\bf CNN Egrid (Egrid CNN\textsubscript{ext}).} We replicate the model proposed by \citet{Dat2017} using their source code.\footnote{\url{https://github.com/datienguyen/cnn_coherence}} The authors generate entity-grid representations of texts (i.e., matrices of entities as columns and sentences as rows, where entities are represented by their syntactic role: subject, object, or other) using the Brown coherence toolkit.\footnote{\url{https://bitbucket.org/melsner/browncoherence}} They then employ a CNN over the entity transitions across sentences in order to capture high-level features and long-range transitions. Training is performed in a pairwise fashion where the model learns to rank a coherent document higher than its incoherent counterparts. To further improve performance, they extend the model by including three entity-specific features, attached to entities' distributed representations: named entity type, salience (represented as the occurrence frequency of entities) and a binary feature indicating whether the entity has a proper mention. 
\vspace{0.2cm} \\ 
\noindent{\bf Lexicalized CNN Egrid (Egrid CNN\textsubscript{lex}).}
The aforementioned Egrid CNN model is agnostic to entities' lexical properties, which are useful features for the task. To remedy this, \citet{Joty2018} further extend it with lexical information about the entities: they represent each entity with its lexical presentation and attach it to its syntactic role (S, O, X). For instance, if ``Obama'' appears as a subject and an object, there will be two different representations for it in the input embedding matrix: Obama-S and Obama-O. 
\citet{Joty2018} achieve state-of-the-art results on the WSJ, outperforming Egrid CNN\textsubscript{ext} without including the three entity-specific features in their model. We also replicate their model using the authors' source code.\footnote{\url{https://ntunlpsg.github.io/project/coherence/n-coh-acl18/}}  
\vspace{0.2cm} \\ 
\noindent{\bf Local Coherence Model (LC).} This model, initially proposed by \citet{Li2014}, applies a window approach to assess a text's local coherence. Sentences are encoded with a recurrent or recursive layer and a filter of weights is applied over each window of sentence vectors to extract ``clique'' scores that are aggregated to calculate the overall document coherence score. We use an improved variant that captures sentence representations via an LSTM and predicts an overall coherence score by averaging the local clique scores \cite{Li2017,Farag2018}. 
\citet{Lai2018} recently showed that the LC model achieves state-of-the-art results on the Clinton and Enron datasets. 
\vspace{0.2cm} \\ 
\noindent{\bf Paragraph sequence (PARSEQ).} \citet{Lai2018} implemented a hierarchical neural network consisting of three LSTMs to generate sentence, paragraph and document representations. 
The network's architecture is similar to our STL model; the key difference is the attention mechanism we use for aggregation. The model was tested on the GCDC and was found to outperform other feature-engineered methods and give state-of-the-art results on the Yahoo dataset.
\vspace{0.2cm} \\ 
\noindent{\bf Neural Single-Task Learning (STL).} We implement the STL model as described in \ref{lex}. For the WSJ data, the network utilizes two Bi-LSTMs to compose sentence and document representations. For the GCDC, we add a third Bi-LSTM, where sentence representations are aggregated via attention to form paragraph vectors. Given these paragraph vectors, we then apply a Bi-LSTM followed by attention to compose the document vectors that are to be scored for coherence.
\vspace{0.2cm} \\
\noindent{\bf Neural Multi-Task Learning (MTL).} We implement the MTL model as described in \ref{mtl}. The same architecture variants as the STL ones are applied on the different datasets.
\vspace{0.2cm} \\
\noindent{\bf Neural S-O-X Multi-Task Learning (MTL\textsubscript{SOX}).} As discussed in \ref{mtl}, we create another version of the MTL model where, for each word, we only predict subject (S), object (O) and `other' (X) roles. 
\vspace{0.2cm} \\
\noindent{\bf GR types Concatenation Model (Concat\textsubscript{grs})}. Instead of learning to predict the GR types within a MTL framework, we incorporate them as input features to the model by concatenating them to the word representations in the STL framework. In this setup, we randomly initialize the types embedding matrix $E_{gr} \in \mathbb{R}^{q \times g}$, where $g$ is the embedding size and $q$ is the number of GR types in the training data. Each type is then mapped to a row in $E_{gr}$ and concatenated to its corresponding word at the model's input layer. Here, the GRs are needed as input at both training and test time, unlike the MTL framework that only requires them during training. The concat\textsubscript{grs} model allows us to further assess whether the MTL framework has an advantage over feeding the GR types as input features.

\begin{table}[]
\centering
\scalebox{0.85}{
\begin{tabular}{|c|c|c|c|c|c|c|}
\hline
\multirow{2}{*}{} & \multirow{2}{*}{\begin{tabular}[c]{@{}c@{}}word embed\\ dim\end{tabular}} & \multicolumn{3}{c|}{LSTM hidden dim} & \multirow{2}{*}{$\alpha$} & \multirow{2}{*}{$\beta$} \\ \cline{3-5}
                  &                                                                       & $h^w$        & $h^s$        & $h^p$        &                        &                       \\ \hline
WSJ               & 50                                                                    & 100        & 100        & -          & 0.7                    & 0.3                   \\ \hline
Yahoo             & 300                                                                   & 100        & 100        & 100        & 1                      & 0.1                   \\ \hline
Clinton           & 300                                                                   & 100        & 200        & 100        & 1                      & 0.1                   \\ \hline
Enron             & 300                                                                   & 100        & 100        & 100        & 1                      & 0.2                   \\ \hline
\end{tabular}
}
\caption{Model hypermarameters: $w$, $s$ and $p$ refer to word, sentence and paragraph hidden layers respectively; $\alpha$ is the main and $\beta$ the secondary loss weight.}
\label{table3}
\end{table}

\subsection{Experimental setup}
\label{expsetup}
We extract the GR types of words using the Stanford Dependency Parser (v. $3.8$) \cite{Chen2014} and obtain a total of $39$ different types of Universal Dependencies and their subtypes (see Appendix A for the full list). 
For the MTL\textsubscript{SOX} model, we consider direct objects, indirect objects and subjects of passive verbs as objects (O). Our models are initialized with pre-trained GloVe embeddings \cite{pennington2014glove}. We use minibatches of size $32$, optimize the models using RMSProp \cite{RMSProp}, and set the learning rate to $0.001$. Dropout \cite{srivastava2014dropout} is used for regularization with probability $0.5$ and applied to the word embedding layer and the output of the Bi-LSTM sentence layer. Table \ref{table3} shows the different hyperparameters used for training.\footnote{We note that hyperparameters are tuned per domain.} 
 
Training is done for $30$ epochs and performance is monitored over the development set; the model with the highest performance (highest PRA on the synthetic data and highest classification accuracy on GCDC) on the development set is selected and applied at testing time. To reduce model variance, we run the WSJ experiments $5$ times with different random initializations and the GCDC ones $10$ times (following \citet{Lai2018}), and average the predicted scores of the ensembles for the final evaluation. For the WSJ data, we use the same train/dev splits as \citet{Dat2017}, and for GCDC, we follow \citet{Lai2018} and split the training data with a $9$:$1$ ratio for tuning.

\section{Results and Discussion}

\noindent{\bf Binary Classification.}
Table \ref{table4} shows the results of the binary discrimination task on the WSJ. The results demonstrate the effectiveness of our MTL approach using a supervision signal at the bottom layers based on the words' GR types, which significantly outperforms all other approaches and achieves state-of-the-art results on the WSJ ($0.932$ PRA and $0.941$ TPRA).\footnote{Significance is calculated based on the randomization test \cite{yeh2000more}.} The performance of the Egrid neural models shows that despite their ability to rank a document higher than its incoherent counterparts ($0.876$ and $0.846$ PRA), they do not generalize when documents are compared against counterparts from the whole test set ($0.656$ and $0.566$ TPRA). This could be partly attributed to the pairwise training strategy adopted by these models and their inability to compare entity-transition patterns across different topics. 
The table also shows that models that utilize compositions over textual units to form document representations (the last four models) are significantly more effective than those explicitly utilizing only the local transitions between sentences (LC model). Furthermore, we observe that incorporating GR types (MTL, MTL\textsubscript{SOX} and Concat\textsubscript{grs}) gives significantly better results compared to the STL model that is GR-agnostic. 
The superiority of the MTL model over Concat\textsubscript{grs} and MTL\textsubscript{SOX} demonstrates that learning the GR types, within an MTL framework, allows the model to learn richer contextual representations (but also to be more efficient at testing time compared to e.g., Concat\textsubscript{grs} since it does not require external linguistic tools).

\begin{table}[t!]
\centering
\scalebox{1.0}{
\begin{tabular}{|c|c|c|}
\hline
Model                                      & PRA            & TPRA           \\ \hline
Egrid CNN\textsubscript{ext}    & 0.876          & 0.656          \\ \hline
Egrid CNN\textsubscript{lex}    & 0.846          & 0.566          \\ \hline
LC                                         & 0.741          & 0.728          \\ \hline
STL                                        & 0.877          & 0.893          \\ \hline
MTL                                        & \textbf{0.932\textsuperscript{*}} & \textbf{0.941\textsuperscript{*}} \\ \hline
MTL\textsubscript{SOX}    & 0.899          & 0.913          \\ \hline
Concat\textsubscript{grs} & 0.896          & 0.908          \\ \hline
\end{tabular}
}
\caption{Results of the binary discrimination task on the WSJ. {\bf*} indicates significance ($p < 0.01$) over all the other models based on the randomization test. Egrid models are significantly worse than MTL\textsubscript{SOX} and Concat\textsubscript{grs} on the PRA metric and significantly worse than all models on TPRA.\footnotemark} 
\label{table4}
\end{table}
\footnotetext{\citet{Joty2018} reported $0.885$ PRA for their Egrid CNN\textsubscript{lex}, which we were unable to replicate using their code; however, this is still lower compared to our results.}

\begin{figure}[tbp]
\centering
\includegraphics[scale=0.4]{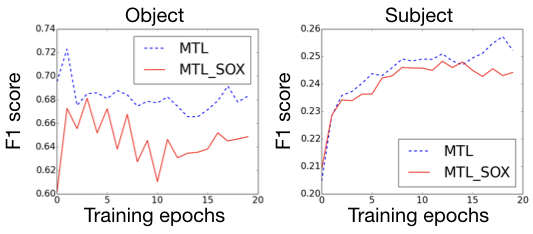}
\caption{F1 scores for subject and object predictions with the MTL and MTL\textsubscript{SOX} models over the first $20$ epochs of training. Y-axis: F1 scores; x-axis: epochs. The graphs are based on the WSJ dev set.} 
\label{figure2}
\end{figure}

To further analyze performance, we calculate the Pearson correlation between: a) the similarity between a permuted document and its original counterpart in terms of the minimum number of adjacent transpositions needed to transform the former back to its original version \cite{Lapata:2006}, and b) the predicted coherence score for the permuted document. This allows us to investigate whether a higher similarity is linked to a higher coherence score. We observe that MTL, MTL\textsubscript{SOX}, Concat\textsubscript{grs} and STL have the highest correlations ($0.260$, $0.232$, $0.227$, $0.225$ respectively), followed by LC ($0.076$), Egrid CNN\textsubscript{ext} ($-0.0126$) and Egrid CNN\textsubscript{lex} ($-0.069$).\footnote{We note that the low correlation is due to the nature of the task: binary evaluation rather than absolute scoring of coherence. }
In order to further analyze the strengths of MTL, we plot in Figure \ref{figure2} the F1 scores over the training epochs for predicting the subject and object types using MTL or MTL\textsubscript{SOX}. We can see that learning to predict a larger set of GR types enhances the model's predictive power for the subject and object types, corroborating the value of entity-based properties for coherence.     

\begin{table}[t!]
\centering
\scalebox{1.0}{
\begin{tabular}{|c|c|c|c|}
\hline
Model       & Yahoo          & Clinton        & Enron          \\ \hline
LC     & 0.535          & 0.610          & 0.544          \\ \hline
PARSEQ & 0.549          & 0.602          & 0.532          \\ \hline
STL    & 0.550          & 0.590          & 0.505          \\ \hline
MTL    & \textbf{0.560} & \textbf{0.620\textsuperscript{*}} & \textbf{0.560\textsuperscript{*}} \\ \hline
MTL\textsubscript{SOX} & 0.505 & 0.585 & 0.510\\ \hline
Concat\textsubscript{grs} & 0.455 & 0.570 & 0.460 \\ \hline
\end{tabular}}
\caption{Model accuracy on the three-way classification task on GCDC. {\bf*} indicates significance over STL with $p < 0.01$ using the randomization test. Results for PARSEQ and LC are those reported in \citet{Lai2018} on the same data. } 
\label{table5}
\end{table}

%Figure 2
%%%%%%%%%%%%%%%%%%%%%%%%%%%%%%%
%%%%%%%%%%%%%%%%%%%%%%%%%%%%%%%

%Figure 3
%%%%%%%%%%%%%%%%%%%%%%%%%%%%%%%
\begin{figure*}[tbp]
\centering
\includegraphics[scale=0.45]{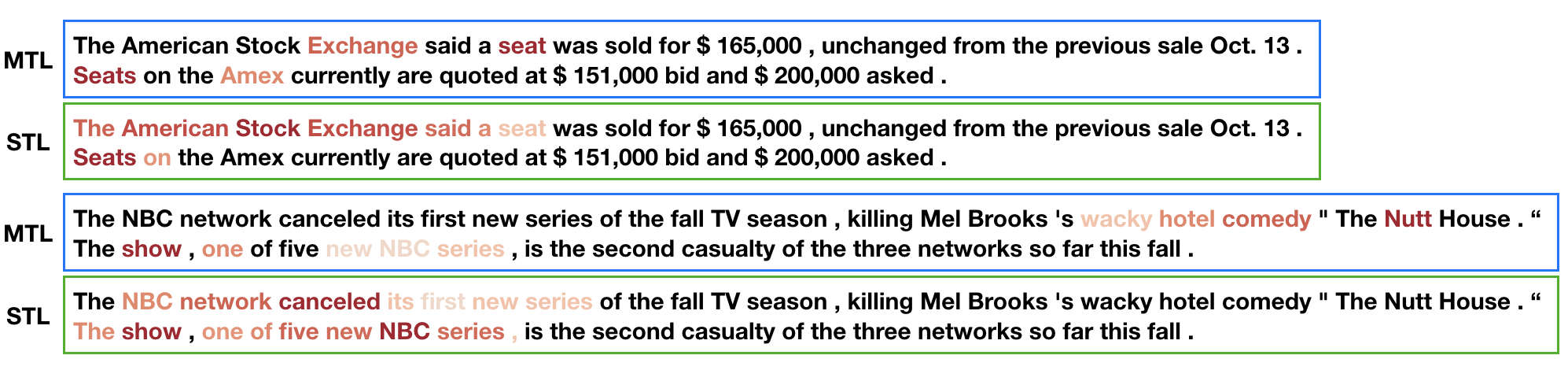}
\caption{Visualization of the model's gradients with respect to the input word embeddings for MTL and STL on the WSJ dev set. Words that contribute the most to coherence scoring (i.e., those with high gradient norms) are colored: the contribution of words decreases from dark red to lighter tones of orange. 
}
\label{figure3}
\end{figure*}
%%%%%%%%%%%%%%%%%%%%%%%%%%%%%%%
\vspace{0.2cm} 
\noindent{\bf Three-way Classification.}
On GCDC (Table \ref{table5}) we can see that MTL achieves state-of-the-art performance across all three datasets. Although different evaluation metrics are employed, we note that the numbers obtained on this dataset are quite low compared to those on the WSJ. Assessing varying degrees of coherence is a more challenging task: differences in coherence between different documents is less pronounced than when taking a document and randomly shuffling its sentences.
When comparing MTL to STL, the former is consistently better across all datasets, with significant improvements for two of them.\footnote{We also note that GR prediction is only required during training; therefore, at inference time, MTL uses the same number of parameters as STL.} 
Interestingly, we observe that MTL\textsubscript{SOX} and Concat\textsubscript{grs} do not generalize to the more realistic domain. 
As shown in Table \ref{table3}, our best MTL model uses smaller $\beta$ and higher $\alpha$ values on the GCDC compared to the WSJ. 
This could be attributed to the performance of the parser and/or the nature of the GCDC and the properties of (in)coherence it exhibits, compared to the WSJ data. 
MTL allows the model more flexibility and control with respect to the features it learns in order to enhance  performance on the main task, in contrast to Concat\textsubscript{grs} where the GRs are given directly as input to the model (yielding the worst performance across all the GCDC datasets). 

The results on GCDC demonstrate that our main MTL approach generalizes to tasks involving the prediction of varying degrees of coherence in everyday writing. In general, however, we observe that, out of the three gold coherence labels (low, medium, high) both MTL and STL have difficulty in correctly classifying documents of medium coherence, which can be attributed to the smaller number of training examples for that class (Section \ref{rwd}). 
\vspace{0.2cm} \\
\noindent{\bf Visualization.}
In an attempt to better understand what the models have learnt, we visualize the words that contribute the most to coherence prediction. We calculate the model's gradients with respect to the input word embeddings (similarly to \citet{Li2015}) to determine which words maximize the model's prediction (more influential words should have higher gradient norms).  
Figure~\ref{figure3} presents example visualizations obtained with STL and MTL.   
We observe that for MTL, important words are those that are considered the center of attention: in the first example (top two sentences) where the document is about seats in the stock exchange, ``seat'' and ``Seats'' are considered more important than the subject entities. On the other hand, the STL model considers the subject of the first sentence (``The American Stock Exchange'') more important than the object ``seat''. In the second example (last two sentences) where the document is about a canceled show by the NBC, for the MTL model, the name of the show (or part of it) in the first sentence (``Nutt'') is considered important, as well as ``comedy'' which also refers to the show; in addition to ``show'' in the second sentence. On the other hand, STL fails to identify the name of the show as important. In general, STL seems to be more distracted, focusing on words that do not necessarily contribute to coherence (e.g., determiners and prepositions), whereas MTL seems to be considering more informative parts of the text.    
\vspace{0.2cm} \\
\noindent{\bf Qualitative Analysis.} 
Following previous work \cite{miltsakaki2004evaluation,Li2017}, we perform a small-scale qualitative analysis: we apply our best model to a number of discourses that exhibit different types of coherence and investigate the predicted coherence scores. We observe that MTL can capture some aspects of lexical and centering/referential coherence: 
\vspace{0.1cm} 

\noindent\textit{Mary ate some apples. She likes apples.} \textbf{0.790}\\
\textit{Mary ate some apples. She likes pears. } \textbf{0.720}\\
\textit{Mary ate some apples. She likes Paris. } \textbf{0.742}\\
\textit{She ate some apples. Mary likes apples. } \textbf{0.747} 
\vspace{0.2cm}

\noindent\textit{John went to his favorite music store to buy a piano. He had frequented the store for many years. } \textbf{0.753} \\
\noindent\textit{John went to his favorite music store to buy a piano. It was a store John had frequented for many years. } \textbf{0.743} 

\vspace{0.1cm} 
On the other hand, it is not as good at recognizing temporal order and causal relationships; for example:

\vspace{0.1cm} 

\noindent\textit{Bret enjoys video games; therefore, he sometimes is late to appointments. } \textbf{0.491}\\
\textit{Bret sometimes is late to appointments; therefore, he enjoys video games.} \textbf{0.499}

\section{Conclusion}
We have presented a hierarchical multi-task learning framework for discourse coherence that takes advantage of inductive transfer between two tasks: predicting the GR type of words at the bottom layers of the network and predicting a document-level coherence score at the top layers. We assessed the extent to which our framework generalizes to different domains and prediction tasks, and demonstrated its effectiveness against a number of baselines not only on standard binary evaluation coherence tasks, but also on tasks involving the prediction of varying degrees of coherence, achieving a new state of the art. As part of future work, we would like to investigate the use of contextualized embeddings (e.g., BERT, \newcite{devlin2018bert}) for coherence assessment -- as such representations have been shown to carry syntactic information of words \cite{tenney2019you} -- and whether they allow multi-task learning frameworks to learn complementary aspects of language. % features to the ones captured by incorporating grammatical roles to our multi-task approach.

\section*{Acknowledgments}
We thank Ted Briscoe and Marek Rei for their valuable suggestions and feedback. We also thank Paula Buttery, Andrew Caines, James Thorne, Christopher Bryant, Simone Teufel and the anonymous ACL reviewers for their insightful comments. We thank the NVIDIA Corporation for the donation of the Titan X Pascal GPU used in this research. We gratefully acknowledge our funding bodies: Youmna Farag was supported by the EPSRC and Cambridge Trust; Helen Yannakoudakis was supported by Cambridge Assessment, University of Cambridge. 

\bibliography{acl2019_v2}

\begin{thebibliography}{37}
\expandafter\ifx\csname natexlab\endcsname\relax\def\natexlab#1{#1}\fi

\bibitem[{Barzilay et~al.(2002)Barzilay, Elhadad, and McKeown}]{Barzilay2002}
Regina Barzilay, Noemie Elhadad, and Kathleen~R. McKeown. 2002.
\newblock Inferring strategies for sentence ordering in multidocument news
  summarization.
\newblock \emph{J. Artif. Int. Res.}, 17(1):35--55.

\bibitem[{Barzilay and Lapata(2005)}]{Barzilay:2005}
Regina Barzilay and Mirella Lapata. 2005.
\newblock Modeling local coherence: An entity-based approach.
\newblock In \emph{Proceedings of the 43rd Annual Meeting on Association for
  Computational Linguistics}, pages 141--148. Association for Computational
  Linguistics.

\bibitem[{Barzilay and Lapata(2008)}]{Barzilay2008}
Regina Barzilay and Mirella Lapata. 2008.
\newblock Modeling local coherence: An entity-based approach.
\newblock \emph{Computational Linguistics}, 3(1):1--34.

\bibitem[{Burstein et~al.(2010)Burstein, Tetreault, and
  Andreyev}]{Burstein2010}
Jill Burstein, Joel Tetreault, and Slava Andreyev. 2010.
\newblock Using entity-based features to model coherence in student essays.
\newblock In \emph{Human Language Technologies: The 2010 Annual Conference of
  the North American Chapter of the Association for Computational Linguistics},
  pages 681--684. Association for Computational Linguistics.

\bibitem[{Chen and Manning(2014)}]{Chen2014}
Danqi Chen and Christopher Manning. 2014.
\newblock A fast and accurate dependency parser using neural networks.
\newblock In \emph{Proceedings of the 2014 Conference on Empirical Methods in
  Natural Language Processing (EMNLP)}, pages 740--750. Association for
  Computational Linguistics.

\bibitem[{Cui et~al.(2018)Cui, Li, Chen, and Zhang}]{Cui2018}
Baiyun Cui, Yingming Li, Ming Chen, and Zhongfei Zhang. 2018.
\newblock Deep attentive sentence ordering network.
\newblock In \emph{Proceedings of the 2018 Conference on Empirical Methods in
  Natural Language Processing}, pages 4340--4349. Association for Computational
  Linguistics.

\bibitem[{Desai et~al.(2018)Desai, Dakle, and Moldovan}]{desai2018generating}
Takshak Desai, Parag Dakle, and Dan Moldovan. 2018.
\newblock Generating questions for reading comprehension using coherence
  relations.
\newblock In \emph{Proceedings of the 5th Workshop on Natural Language
  Processing Techniques for Educational Applications}, pages 1--10.

\bibitem[{Devlin et~al.(2018)Devlin, Chang, Lee, and
  Toutanova}]{devlin2018bert}
Jacob Devlin, Ming-Wei Chang, Kenton Lee, and Kristina Toutanova. 2018.
\newblock Bert: Pre-training of deep bidirectional transformers for language
  understanding.
\newblock \emph{arXiv preprint arXiv:1810.04805}.

\bibitem[{Elsner and Charniak(2008)}]{Elsner2008}
Micha Elsner and Eugene Charniak. 2008.
\newblock Coreference-inspired coherence modeling.
\newblock In \emph{Proceedings of ACL-08: HLT, Short Papers}, pages 41--44.
  Association for Computational Linguistics.

\bibitem[{Elsner and Charniak(2011)}]{Elsner2011}
Micha Elsner and Eugene Charniak. 2011.
\newblock Extending the entity grid with entity-specific features.
\newblock In \emph{Proceedings of the 49th Annual Meeting of the Association
  for Computational Linguistics: Human Language Technologies}, pages 125--129.
  Association for Computational Linguistics.

\bibitem[{Farag et~al.(2018)Farag, Yannakoudakis, and Briscoe}]{Farag2018}
Youmna Farag, Helen Yannakoudakis, and Ted Briscoe. 2018.
\newblock Neural automated essay scoring and coherence modeling for
  adversarially crafted input.
\newblock In \emph{Proceedings of the 2018 Conference of the North American
  Chapter of the Association for Computational Linguistics: Human Language
  Technologies, Volume 1 (Long Papers)}, pages 263--271. Association for
  Computational Linguistics.

\bibitem[{Feng et~al.(2014)Feng, Lin, and Hirst}]{Feng2014}
Vanessa~Wei Feng, Ziheng Lin, and Graeme Hirst. 2014.
\newblock The impact of deep hierarchical discourse structures in the
  evaluation of text coherence.
\newblock In \emph{Proceedings of COLING 2014, the 25th International
  Conference on Computational Linguistics: Technical Papers}, pages 940--949.
  Dublin City University and Association for Computational Linguistics.

\bibitem[{Grosz et~al.(1995)Grosz, Weinstein, and Joshi}]{Grosz1995}
Barbara~J. Grosz, Scott Weinstein, and Aravind~K. Joshi. 1995.
\newblock Centering: A framework for modeling the local coherence of discourse.
\newblock \emph{Computational Linguistics}, 21(2).

\bibitem[{Guinaudeau and Strube(2013)}]{Guinaudeau2013}
Camille Guinaudeau and Michael Strube. 2013.
\newblock Graph-based local coherence modeling.
\newblock In \emph{Proceedings of the 51st Annual Meeting of the Association
  for Computational Linguistics (Volume 1: Long Papers)}, pages 93--103.
  Association for Computational Linguistics.

\bibitem[{Hochreiter and Schmidhuber(1997)}]{Hochreiter1997}
Sepp Hochreiter and J\"{u}rgen Schmidhuber. 1997.
\newblock Long short-term memory.
\newblock \emph{Neural Comput.}, 9(8):1735--1780.

\bibitem[{Joty et~al.(2018)Joty, Mohiuddin, and Tien~Nguyen}]{Joty2018}
Shafiq Joty, Muhammad~Tasnim Mohiuddin, and Dat Tien~Nguyen. 2018.
\newblock Coherence modeling of asynchronous conversations: A neural entity
  grid approach.
\newblock In \emph{Proceedings of the 56th Annual Meeting of the Association
  for Computational Linguistics (Volume 1: Long Papers)}, pages 558--568.
  Association for Computational Linguistics.

\bibitem[{Lai and Tetreault(2018)}]{Lai2018}
Alice Lai and Joel Tetreault. 2018.
\newblock Discourse coherence in the wild: A dataset, evaluation and methods.
\newblock In \emph{Proceedings of the 19th Annual SIGdial Meeting on Discourse
  and Dialogue}, pages 214--223. Association for Computational Linguistics.

\bibitem[{Lapata(2006)}]{Lapata:2006}
Mirella Lapata. 2006.
\newblock Automatic evaluation of information ordering: Kendall's tau.
\newblock \emph{Comput. Linguist.}, 32(4):471--484.

\bibitem[{Li et~al.(2016)Li, Chen, Hovy, and Jurafsky}]{Li2015}
Jiwei Li, Xinlei Chen, Eduard Hovy, and Dan Jurafsky. 2016.
\newblock Visualizing and understanding neural models in {NLP}.
\newblock In \emph{Proceedings of the 2016 Conference of the North American
  Chapter of the Association for Computational Linguistics: Human Language
  Technologies}, pages 681--691. Association for Computational Linguistics.

\bibitem[{Li and Hovy(2014)}]{Li2014}
Jiwei Li and Eduard Hovy. 2014.
\newblock A model of coherence based on distributed sentence representation.
\newblock In \emph{Proceedings of the 2014 Conference on Empirical Methods in
  Natural Language Processing (EMNLP)}, pages 2039--2048. Association for
  Computational Linguistics.

\bibitem[{Li and Jurafsky(2017)}]{Li2017}
Jiwei Li and Dan Jurafsky. 2017.
\newblock Neural net models of open-domain discourse coherence.
\newblock In \emph{Proceedings of the 2017 Conference on Empirical Methods in
  Natural Language Processing}, pages 198--209. Association for Computational
  Linguistics.

\bibitem[{Lin et~al.(2011)Lin, Ng, and Kan}]{Lin2011}
Ziheng Lin, Hwee~Tou Ng, and Min-Yen Kan. 2011.
\newblock Automatically evaluating text coherence using discourse relations.
\newblock In \emph{Proceedings of the 49th Annual Meeting of the Association
  for Computational Linguistics: Human Language Technologies}, pages 997--1006.
  Association for Computational Linguistics.

\bibitem[{Logeswaran et~al.(2018)Logeswaran, Lee, and Radev}]{Logeswaran2018}
Lajanugen Logeswaran, Honglak Lee, and Dragomir~R. Radev. 2018.
\newblock Sentence ordering and coherence modeling using recurrent neural
  networks.
\newblock In \emph{AAAI}, pages 5285--5292. AAAI Press.

\bibitem[{Louis and Nenkova(2012)}]{Louis2012}
Annie Louis and Ani Nenkova. 2012.
\newblock A coherence model based on syntactic patterns.
\newblock In \emph{Proceedings of the 2012 Joint Conference on Empirical
  Methods in Natural Language Processing and Computational Natural Language
  Learning}, pages 1157--1168. Association for Computational Linguistics.

\bibitem[{Miltsakaki and Kukich(2004)}]{miltsakaki2004evaluation}
Eleni Miltsakaki and Karen Kukich. 2004.
\newblock Evaluation of text coherence for electronic essay scoring systems.
\newblock \emph{Natural Language Engineering}, 10(1):25--55.

\bibitem[{Parveen et~al.(2016)Parveen, Mesgar, and Strube}]{Daraksha2016}
Daraksha Parveen, Mohsen Mesgar, and Michael Strube. 2016.
\newblock Generating coherent summaries of scientific articles using coherence
  patterns.
\newblock In \emph{Proceedings of the 2016 Conference on Empirical Methods in
  Natural Language Processing}, pages 772--783. Association for Computational
  Linguistics.

\bibitem[{Pennington et~al.(2014)Pennington, Socher, and
  Manning}]{pennington2014glove}
Jeffrey Pennington, Richard Socher, and Christopher~D. Manning. 2014.
\newblock Glove: Global vectors for word representation.
\newblock In \emph{Empirical Methods in Natural Language Processing (EMNLP)},
  pages 1532--1543.

\bibitem[{Rei and Yannakoudakis(2017)}]{rei2017auxiliary}
Marek Rei and Helen Yannakoudakis. 2017.
\newblock Auxiliary objectives for neural error detection models.
\newblock In \emph{Proceedings of the 12th Workshop on Innovative Use of {NLP}
  for Building Educational Applications}.

\bibitem[{Sanh et~al.(2018)Sanh, Wolf, and Ruder}]{sanh2018hierarchical}
Victor Sanh, Thomas Wolf, and Sebastian Ruder. 2018.
\newblock A hierarchical multi-task approach for learning embeddings from
  semantic tasks.
\newblock \emph{arXiv preprint arXiv:1811.06031}.

\bibitem[{Somasundaran et~al.(2014)Somasundaran, Burstein, and
  Chodorow}]{Somasundaran2014}
Swapna Somasundaran, Jill Burstein, and Martin Chodorow. 2014.
\newblock Lexical chaining for measuring discourse coherence quality in
  test-taker essays.
\newblock In \emph{Proceedings of COLING 2014, the 25th International
  Conference on Computational Linguistics: Technical Papers}, pages 950--961.
  Dublin City University and Association for Computational Linguistics.

\bibitem[{Srivastava et~al.(2014)Srivastava, Hinton, Krizhevsky, Sutskever, and
  Salakhutdinov}]{srivastava2014dropout}
Nitish Srivastava, Geoffrey~E Hinton, Alex Krizhevsky, Ilya Sutskever, and
  Ruslan Salakhutdinov. 2014.
\newblock Dropout: a simple way to prevent neural networks from overfitting.
\newblock \emph{Journal of Machine Learning Research}, 15(1):1929--1958.

\bibitem[{Tenney et~al.(2019)Tenney, Xia, Chen, Wang, Poliak, McCoy, Kim,
  Van~Durme, Bowman, Das et~al.}]{tenney2019you}
Ian Tenney, Patrick Xia, Berlin Chen, Alex Wang, Adam Poliak, R~Thomas McCoy,
  Najoung Kim, Benjamin Van~Durme, Samuel~R Bowman, Dipanjan Das, et~al. 2019.
\newblock What do you learn from context? probing for sentence structure in
  contextualized word representations.
\newblock \emph{arXiv preprint arXiv:1905.06316}.

\bibitem[{Tieleman and Hinton(2012)}]{RMSProp}
Tijmen Tieleman and Geoffrey Hinton. 2012.
\newblock Lecture 6.5 - rmsprop.
\newblock \emph{Technical report}.

\bibitem[{Tien~Nguyen and Joty(2017)}]{Dat2017}
Dat Tien~Nguyen and Shafiq Joty. 2017.
\newblock A neural local coherence model.
\newblock In \emph{Proceedings of the 55th Annual Meeting of the Association
  for Computational Linguistics (Volume 1: Long Papers)}, pages 1320--1330.
  Association for Computational Linguistics.

\bibitem[{Verberne et~al.(2007)Verberne, Boves, Oostdijk, and
  Coppen}]{verberne2007evaluating}
Suzan Verberne, LWJ Boves, NHJ Oostdijk, and PAJM Coppen. 2007.
\newblock Evaluating discourse-based answer extraction for why-question
  answering.
\newblock In \emph{Proceedings of the 30th Annual International ACM SIGIR
  Conference on Research and Development in Information Retrieval.}, pages
  735--736.

\bibitem[{Yannakoudakis and Briscoe(2012)}]{y2012modeling}
Helen Yannakoudakis and Ted Briscoe. 2012.
\newblock Modeling coherence in esol learner texts.
\newblock In \emph{Proceedings of the Seventh Workshop on Building Educational
  Applications Using NLP}, pages 33--43. Association for Computational
  Linguistics.

\bibitem[{Yeh(2000)}]{yeh2000more}
Alexander Yeh. 2000.
\newblock More accurate tests for the statistical significance of result
  differences.
\newblock In \emph{Proceedings of the 18th conference on Computational
  linguistics-Volume 2}, pages 947--953. Association for Computational
  Linguistics.

\end{thebibliography}
\bibliographystyle{acl_natbib}

\vfill\eject

\appendix
\section{Grammatical roles}
\begin{table}[h!]
\resizebox{\columnwidth}{!}{\begin{minipage}{\columnwidth}
\centering
\scalebox{0.9}{
\begin{tabular}{|c|c|}
\hline
Type                                                                   & Description                                  \\ \hline
\begin{tabular}[c]{@{}c@{}}acl\\ {[}relcl{]}\end{tabular}              & \begin{tabular}[c]{@{}c@{}}clausal modifier of noun\\ (adjectival clause)\end{tabular} \\ \hline
advcl                                                                  & adverbial clause modifier                    \\ \hline
advmod                                                                 & adverbial modifier                           \\ \hline
amod                                                                   & adjectival modifier                          \\ \hline
appos                                                                  & appositional modifier                        \\ \hline
aux                                                                    & auxiliary                                    \\ \hline
auxpass                                                                & passive auxiliary                            \\ \hline
case                                                                   & case marking                                 \\ \hline
\begin{tabular}[c]{@{}c@{}}cc\\ {[}preconj{]}\end{tabular}             & coordinating conjunction                     \\ \hline
ccomp                                                                  & clausal complement                           \\ \hline
\begin{tabular}[c]{@{}c@{}}compound\\ {[}prt{]}\end{tabular}       & compound                                     \\ \hline
conj                                                                   & conjunct                                     \\ \hline
cop                                                                    & copula                                       \\ \hline
csubj                                                                  & clausal subject                              \\ \hline
csubjpass                                                              & clausal passive subject                      \\ \hline
dep                                                                    & unspecified dependency                       \\ \hline
\begin{tabular}[c]{@{}c@{}}det\\ {[}predet{]}\end{tabular}             & determiner                                   \\ \hline
discourse                                                              & discourse element                            \\ \hline
dobj                                                                   & direct object                                \\ \hline
expl                                                                   & expletive                                    \\ \hline
iobj                                                                   & indirect object                              \\ \hline
mark                                                                   & marker                                       \\ \hline
mwe                                                                    & multi-word expression                        \\ \hline
neg                                                                    & negation modifier                            \\ \hline
\begin{tabular}[c]{@{}c@{}}nmod\\ {[}tmod, poss, npmod{]}\end{tabular} & nominal modifier                             \\ \hline
nsubj                                                                  & nominal subject                              \\ \hline
nsubjpass                                                              & passive nominal subject                      \\ \hline
nummod                                                                 & numeric modifier                             \\ \hline
parataxis                                                              & parataxis                                    \\ \hline
punct                                                                  & punctuation                                  \\ \hline
root                                                                   & root                                         \\ \hline
xcomp                                                                  & open clausal complement                      \\ \hline
\end{tabular}}
\caption{The GR types (UDs) extracted from the WSJ training data. The text inside [] (left column) denotes the extracted subtypes (language specific types).\footnote{For more details about subtypes please see \url{http://universaldependencies.org/docsv1/ext-dep-index.html}.} The total number of main types and their subtypes is $39$.\footnote{For the full list of UDs please see \url{http://universaldependencies.org/docsv1/u/dep/index.html}.}}
 \end{minipage}}
\end{table}

\end{document}